\documentclass[11pt,a4paper]{article}
\usepackage{hyperref}
\usepackage{acl2021}
\usepackage{times}
\usepackage{latexsym}

\usepackage{microtype}

\aclfinalcopy 

\title{End-to-End Self-Debiasing Framework for Robust NLU Training}

\author{Abbas Ghaddar, Philippe Langlais\textsuperscript{\textdagger}, Mehdi Rezagholizadeh and Ahmad Rashid    \\
	Huawei Noah's Ark Lab, Montreal Research Center, Canada \\
	\textsuperscript{\textdagger}RALI-DIRO, Universit\'e de Montr\'eal,	Montr\'eal, Canada \\
	{\tt abbas.ghaddar@huawei.com, felipe@iro.umontreal.ca}  \\
	{\tt mehdi.rezagholizadeh@huawei.com,
	ahmad.rashid@huawei.com}
	\\}

\newcommand{\bert}{\textsc{Bert}}

\newcommand{\mightmention}[1]{}
\newcommand{\problem}[1]{\textcolor{red}{$\star$}}
\newcommand{\answer}[1]{\textcolor{blue}{$\#$}}
\newcommand{\todoreview}[1]{\textcolor{green}{$@$}}
\newcommand{\real}{\mathbb{R}}

\usepackage{subcaption}
\usepackage{rotating,csquotes}
\usepackage{xcolor}
\usepackage{multirow}
\usepackage{framed}
\usepackage{booktabs}
\usepackage{amsmath}
\usepackage{upgreek}
\usepackage{ulem}
\usepackage{amsfonts}

\usepackage[most]{tcolorbox}

\newtcbox{\mybox}[1][]{enhanced, colframe=blue, colback=blue!15, 
	frame style={opacity=0.25}, interior style={opacity=0.25}, 
	nobeforeafter, tcbox raise base, shrink tight, extrude by=1mm, #1}

\date{}

\begin{document}
\maketitle
\begin{abstract}

Existing Natural Language Understanding (NLU) models have been shown to incorporate dataset biases leading to strong performance on in-distribution (ID) test sets but poor performance on out-of-distribution (OOD) ones. We introduce a simple yet effective debiasing framework whereby the shallow representations of the main model are used to derive a bias model and both models are trained simultaneously. We demonstrate on three well studied NLU tasks that despite its simplicity, our method leads to competitive OOD results. It significantly outperforms other debiasing approaches on two tasks, while still delivering high in-distribution performance.
 
\end{abstract}

\section{Introduction}
  
Researchers have increasingly raised concerns about the tendency of recent NLU models~\cite{devlin2018bert,liu2019roberta} to quickly leverage spurious surface lexical-syntactic features~\cite{poliak2018hypothesis,gururangan2018annotation,dasgupta2018evaluating,ghaddar2017winer}. These superficial properties, also referred as dataset biases~\cite{shah2020predictive,utama2020towards,moosavi2020improving}, result in significant performance drop on out-of-distribution (OOD) sets containing counterexamples to biases in the training data~\cite{mccoy2019right,schuster2019towards,zhang2019paws,ghaddar2021context}.

The most common approach to tackle the problem consists in training a \textit{bias model} with hand-crafted features with the goal of identifying biased training examples. This information is used in a later stage to discourage the \textit{main model} from adopting the  naive strategy of the bias model. Several debiasing training paradigms have been proposed to adjust the importance of biased training samples, such as  product of experts~\cite{clark2019don,he2019unlearn}, learned-mixin~\cite{clark2019don}, example reweighting~\cite{schuster2019towards}, debiased focal loss~\cite{mahabadi2020end}, and confidence regularization~\cite{utama2020mind}. 

Recently, there has been a number of endeavours to produce a bias model without prior knowledge on the targeted biases or without the need for manually designing features. \citet{utama2020towards} propose to use instead a model trained on a tiny fraction ($<1\%$) of the training data for few epochs as a bias model; while \citet{clark2020learning}~and \citet{sanh2020learning} trained a low capacity model on the full training set. These approaches target the training of the bias model alone, which is subsequently queried while training the main model of interest.

In this paper, we propose an end-to-end debiasing framework which does not require an extra training stage, or manual  bias features engineering. The bias model is indeed a simple attention-based classification layer on top of the main model's intermediate representations. Both models are trained simultaneously in an end-to-end manner as in \cite{mahabadi2020end}, where the importance of training samples for both models are adjusted using the example reweighting technique of~\cite{schuster2019towards}.\footnote{The bias model for these two works is based on hand-crafted features.} The idea of using intermediate classifiers has previously been explored to reduce the inference cost~\cite{schwartz2020right,zhou2020bert,xin2021berxit} for Transformer-based~\cite{vaswani2017attention} models.

In contrast to all previous works, our bias model helps locating lexico-syntactic bias features, inside the main model's intermediate layers, whose importance is reduced by adding a noise vector; therefore preventing the main model to rely on them. During training, both the main and bias models interact by interchangeably re-weighting the importance of each others' examples. 
 
Our learning framework, when applied to a vanilla \bert{}~\cite{devlin2018bert} model leads to consistent and significant improvements on 3 NLU tasks, while maintaining a balanced performance between ID and OOD sets. It involves a single training stage, and only incurs a small number of extra parameters (0.5M) compared to other approaches. For instance, \citet{utama2020mind} used a copy of the main model (110M parameters)  as the bias model, while \cite{sanh2020learning} used \bert{-Tiny}~\cite{turc2019well} that has 11M parameters and has been pre-trained from scratch using the masked LM objective~\cite{devlin2018bert}.

\section{Method}

\begin{figure}[!ht]
 	\begin{center}
 			\includegraphics[scale=1.1]{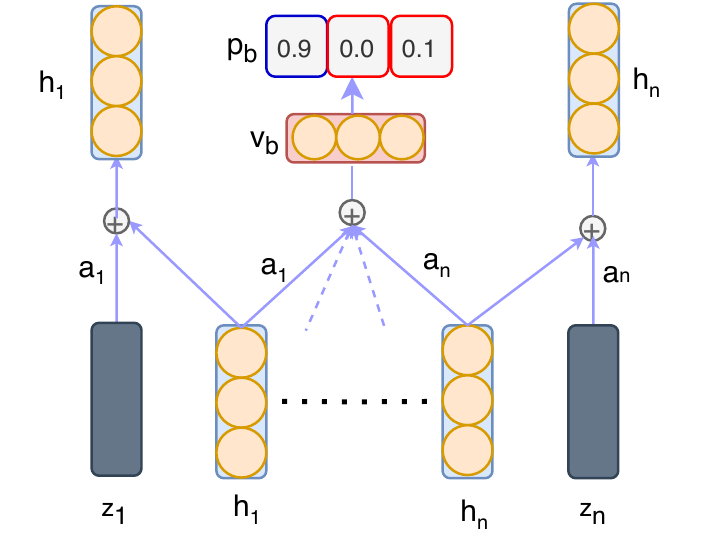}
 	\end{center}	
 	\caption{Illustration of our framework where the bias model is placed at the top of the $k^{th}$ layer of the main model. Notation is introduced in the text, for simplicity we use $h_i$ to refer to $h^k_i$. The blue rectangle indicates the index of of the ground truth class.}
 	\label{fig:bias_model}
	
 \end{figure}

We describe an end-to-end solution for debiasing a Transformer-based  classification model. We note $x=\{x_1, x_2,\ldots, x_n\}$ the input sequence of length $n$, and $y \in \{1, 2, \ldots, T\}$ the gold label, where $T$ is the number of classes. Figure~\ref{fig:bias_model} shows a diagram of our method.

\subsection{Main Model}
\label{sec:Main Model}

The main model is a Transformer-based \bert{} encoder~\cite{vaswani2017attention,devlin2018bert} with a classifier on top of the classification \texttt{[CLS]} token of the last layer. This classifier generates the probability distribution $p_m \in \real^{T}$ over the output classes given an input $x$. At each layer $k$, \bert{} produces an internal hidden representation $h^k_i \in \real^{d}$  for each token $x_i$ in the input. 
 
\subsection{Bias Model}
\label{sec:Bias Model}

We hypothesize that the bottom layers of the main model can serve as input for the bias model, thus avoiding the need for an external model or for manually designed features. Our intuition is built on the observation made by~\citet{jawahar2019does} that bottom layers of \bert{} mainly encode lexical and syntactic information, and on the fact that such models tend to quickly overfit this type of information \cite{zellers2018swag,mccoy2019right}.

Our  bias model is composed of an additive attention layer~\cite{bahdanau2014neural}, followed by a softmax one. This module uses the $k^{th}$ layer of the main model as input in order to produce a probability distribution over classes. 
First, a scalar value $\tilde{a}_{i} \in \real$ is computed using a feed forward neural network with weight matrices $W_e \in \real^{ d \times d}$ and $W_a \in \real^{ 1 \times d}$ such that:
\begin{equation}
    \tilde{a}_{i} = W_a\tanh \left( W_e h^k_i \right)  
\end{equation}

Those scalar values are normalized and referred to as attentions $a_{i}$:
\begin{equation}
    a_{i} = \frac{\exp(\tilde{a}_{i})}{\sum_{i=1}^n \exp(\tilde{a}_{i})}
    \label{attention}
\end{equation}

The representation vector, $v_{b}$, for the bias model is computed as the weighted sum of the intermediate representations of the $k^{th}$ layer:
\begin{equation}
    v_{b} = \sum_{i=1}^{n} a_{i} h^k_i
\end{equation}
We obtain the probability  distribution $p_b$ over the classes by applying a projection layer (parameterized by $W_{y} \in \real^{ d \times T}$) followed by a softmax function: $p_{b} = \text{softmax}(W_{y}^{\top}v_{b})$.

\subsection{Biased Features Regularization}
\label{sec:Removing Bias Features}
 
The attention weights of the bias model reveal which hidden representations are the most informative for the classification decision of that model. We propose to de-emphasize these ``bias features'' from the main model. We do so by adding a weighted noise vector to the hidden representations of the main model at layer $k$. Let $z \leftarrow \{z_1,\ldots,z_n\}$  be a set of zero-mean Gaussian noise vectors corresponding to each token in the sequence, where $z_i \in \real^d$. The intermediate  representations of the main model at layer $k$ are updated as follows:
 \begin{equation}
    h_{i}^{k} \leftarrow a_i z_i + h_{i}^{k}
\end{equation}
where $a_i$ determines the relative amplitude of noise added to each token representation. A high value of $a_i$ means that $h_{i}^{k}$ most likely contains bias features. Consequently, a large amount of noise is added at this position, which will prevent the main model from overfitting to these features at subsequent layers.

\subsection{Example Re-weighting}
\label{sec:Example Reweighting}

For both models, we adjust the importance of a training sample by directly assigning it a scalar weight that indicates whether the sample exhibits a bias or not. Let $p_m^c$ and $p_b^c$ be the predicted probabilities corresponding to the ground-truth class by the bias and main model respectively. The weight assigned to the main model training sample is calculated as follows: 
\[
w_{m}= 
\begin{cases}
1-p_b^c             &{p_b^c > \gamma}\\
1 &  \text{otherwise} \\
\end{cases}
\]

Differently from the reweighing method of~\cite{clark2019don,schuster2019towards}, we add a hard threshold $\gamma \in [0,1]$ to control the number of samples to be re-weighted. The goal is to ensure that the importance of main model samples will be attenuated only if the bias model predictions fall into the high confidence bin ($\gamma >0.8$) on biased samples. To further strengthen this constraint, we down-weight the importance of bias model training samples if the main model confidence is below a threshold $\beta$:
\[
w_{b}= 
\begin{cases}
p_m^c             &{p_m^c < \beta}\\
1 &  \text{otherwise} \\
\end{cases}
\]

This ensures that the bias model is focusing on easy examples at early training stages, while challenging ones are gradually fed in the later training steps. Since the main and bias models are trained simultaneously, they interchangeably re-weight each others' examples. This is different from previous works where all samples have the same importance during the training phase of the bias model. For a single training instance, the individual loss term for the main and bias models are:
\begin{eqnarray}
    \mathcal{L}(\theta_B) = -w_{b}log(p_b^c)\\
    \mathcal{L}(\theta_M) = -w_{m}log(p_m^c)
\end{eqnarray}

where $\theta_B=\{W_e, W_a, W_{y}\}$ are the trainable parameters for the bias model, while $\theta_{M}$ are the ones for the main model (\bert{} parameters). We train both models by minimizing the aforementioned losses. At inference time, only the main model is used for prediction, and no noise is added to $h^k$. It is worth mentioning that our proposed extension has no impact on the overall training time. 

\begin{table*}[!th]
	\begin{center}
		\begin{tabular}{l|cc|cc|cc}
			\toprule
			\multirow{2}{*}{Model} & \multicolumn{2}{c|}{MNLI} & \multicolumn{2}{c|}{FEVER} &
			\multicolumn{2}{c}{QQP} \\
			 & Dev & HANS & Dev & Symm. & Dev & PAWS \\
			 \midrule
			\bert{-base} & 84.5 & 62.4 & 85.6 & 55.1 & \bf{91.0} & 33.5 \\
			\midrule
		    \cite{clark2019don} & 83.5 & 69.2 & - & - & - & - \\
			\cite{schuster2019towards} & - & - & 85.4 & 61.7 & - & -\\
			\cite{utama2020mind} & \textbf{84.5} & 69.1 & 86.4 & 60.5 & 89.1 & 40.0\\
			\cite{utama2020towards}\textdagger & 82.3 & 69.7 & - & - & 85.2 & \textbf{57.4}\\
			\cite{utama2020towards}$\ddagger$ & 84.3 & 67.1 & - & - & 89.0 & 43.0\\
			
			\cite{sanh2020learning}$\spadesuit$  & 81.4 & 68.8 & 82.0 & 60.0 & -& - \\
			\cite{sanh2020learning}$\clubsuit$  & 83.3 & 67.9 & 85.3 & 57.9 & -& - \\
			\midrule 
		     this work & 83.2$\pm$0.1 & \textbf{71.2}$\pm$0.2 & \textbf{86.9} $\pm$0.8 & \bf{63.8}$\pm$0.3 & 90.2$\pm$0.2 & 46.5$\pm$2.3\\
		     \hspace{3mm}w/o noise & 83.7$\pm$0.3 & 68.6$\pm$0.9 & 85.5$\pm$0.5 & 61.6$\pm$0.6 & 90.4$\pm$0.3 & 42.4$\pm$2.1\\
		     \hspace{3mm}w/o main reweighting & 84.0$\pm$0.4 & 62.8$\pm$1.1 & 85.1$\pm$0.4 & 57.4 $\pm$0.8 & 91.0$\pm$0.2 & 36.7$\pm$1.6 \\
		     \hspace{3mm}w/o bias reweighting & 81.5$\pm$0.9 & 64.6$\pm$0.6 & 83.9$\pm$0.4 & 60.3$\pm$0.5 & 89.3$\pm$0.7 & 39.7$\pm$2.4\\
			
			\bottomrule
		\end{tabular}
	\end{center}
	\caption{Model performance when evaluated on MNLI, Fever, QQP, and their corresponding challenge test sets. Ablation study of our method without adding noise, and without reweighting main or bias model's training samples respectively.  Symbols are used to distinguish variants from the same paper that use a different training technique: (\textdagger) example reweighting, ($\ddagger$) confidence regularization, ($\spadesuit$) product of Experts (PoE), ($\clubsuit$) PoE+ cross-entropy.}
	\label{tab:res}
	
\end{table*}

\section{Experiments}

\subsection{Datasets}

We test our method on 3 sentence-pair classification tasks supported by ID training and validation sets as well as an OOD test set, which are specifically built to measure robustness to dataset bias. 

For the \textbf{Natural Language Inference} task,  we use the MNLI~\cite{williams2017broad} benchmark as our ID data, and  HANS~\cite{mccoy2019right} as our OOD test set. For \textbf{Fact Verification}, we use the FEVER~\cite{thorne2018fever} benchmark for ID evaluation, and FEVER Symmetric~\cite{schuster2019towards} (version 1) as our OOD test set. For \textbf{Paraphrase identification}, we use QQP~\footnote{\url{https://www.quora.com/q/quoradata/First-Quora-Dataset-Release-Question}} as our ID data, and PAWS~\cite{zhang2019paws} as our OOD test set.

\subsection{Implementation}

We use the 12-layer \bert{-base} model~\cite{devlin2018bert} as our main model, thus our results can be compared with prior works. We adopt the standard setup of \bert{} and represent a pair of sentences as: \texttt{[CLS]} 1\textsuperscript{st} sentence \texttt{[SEP]} 2\textsuperscript{nd} sentence \texttt{[SEP]}. For \bert{} hyper-parameters, we use those of the baseline: a batch size of 64, learning rate of 1e-5 with the Adam~\cite{kingma2014adam} optimizer.

Following~\cite{clark2019don,grand2019adversarial,clark2020learning,sanh2020learning}, we tune our method hyperparameters on the OOD sets. As pointed out by~\cite{clark2019don,clark2020learning}, this is not ideal since it assumes some prior knowledge of the OOD test sets. To best mitigate this impact, we follow the procedure of previous works and use the same hyper-parameters for all 3 tasks. We varied $k$ ($[2-5]$), $\gamma$ ($[0.7-0.9]$), $\beta$ ($[0.5-0.7]$), and fix their values to  3, 0.8, 0.5 respectively. 

However, different set of parameters performed roughly equally well, provided that $k$ is 3 or 4, $\gamma$ $\geq$ 0.8 and $\beta$ is set to 0.5. We use  early  stopping and report mean performance and standard deviation over 6 runs with different seeds.

\subsection{Results}

Table~\ref{tab:res} reports accuracy scores on the development and OOD test sets of the 3 benchmarks we considered. The baseline is the vanilla \bert{-base} model which is used as a backbone for the main model in all the configurations reported. Also, the table shows the ablation results of our method without adding noise, and  without reweighting the training instances of the bias and main models (setting $\beta$ and $\gamma$ to 0 or  1 respectively).   

First, we notice high variance in performances between runs in the debiasing setting, which was also reported in~\cite{clark2019don,mahabadi2020end,utama2020towards,sanh2020learning}. Second, we observe that our method offers a good balance between gains on OOD test sets over the baseline, and losses on ID sets. More precisely, we report the best results on FEVER (both ID and OOD test sets), while we improve the HANS score on the MNLI task, but fail to maintain the baseline score on dev as \citet{utama2020mind} did. 

On QQP, \citet{utama2020towards} reported a much higher score on PAWS (57.4\% vs. 46.5\%), but at the expense of an important drop on the ID dev set (85.2\% vs. 90.2\%). However, our method outperforms this particular model on both MNLI sets, and also shows better cross-task performances compared to prior works. The results are satisfactory, especially when considering the simplicity and efficiency of our approach. Moreover, the fact that a single configuration works well on 3 tasks is an indicator that our method has the potential to generalize on completely unknown OOD sets~\cite{clark2020learning}.

Expectedly, deactivating the main model reweighting mechanism results in near baseline performances. Solely adding noise signal leads to a modest gain of 2-3\% on the OOD test sets and a slight drop ($<1\%$) on the dev sets compared to the baseline. On one hand, without adding noise, our scores are comparable with previous works across the 3 tasks, that is, a significant drop on OOD test sets and minor gains on ID ones. These observations suggest that down-weighting biased examples is important, while de-emphasizing bias features further improves robustness.

\begin{table}[!th]
	
	\begin{center}
		
		\begin{tabular}{l|cc|cc}
			\toprule
			\multirow{2}{*}{Task} & \multicolumn{2}{c|}{Acc.} & \multicolumn{2}{c}{$p^c_b>\gamma$}\\ 
			& w & w/o & w & w/o \\
			\midrule
            MNLI & 67.9 & 69.7  & 18\% & 23\% \\ 
            FEVER & 70.8 & 85.7 & 16\% & 28\% \\
            QQP & 87.6 &  92.9 & 30\% & 44\% \\  
			\bottomrule
		\end{tabular}
	\end{center}	
	\caption{Bias model training accuracy and percentage of training samples correctly classified with high probability, with and without reweighting bias model examples.}
	\label{tab:bias}
\end{table}

On the other hand, we observe that not reweighting bias model examples results in the worst performance on the ID sets. We conduct an analysis on the bias model to better understand the impact of reweighting its examples. As shown in Table~\ref{tab:bias}, its training accuracy as well as the percentage of high confidence predictions increases when examples are not re-weighted, further down-weighting the main model's training examples, which leads to a significant drop of both the ID and OOD performances for all 3 tasks.

\begin{figure}[!h]
			
	\begin{tabular}{lp{7cm}}
		
		a) & \textbf{GT:} \textit{C}; \textbf{Main:} 0.81; \textbf{Bias (w/o):} 0.99 (0.94)\\
		& \textbf{H} Yes, sir. \\
		& \textbf{P} No, not in particular Sir. \\
		\\
		b) & \textbf{GT:} \textit{E}; \textbf{Main:} 0.74; \textbf{Bias (w/o):} 0.95 (0.92)\\
		& \textbf{H} right after the war \\
		& \textbf{P} Just after the war ended. \\
		\\
		c) & \textbf{GT:} \textit{C}; \textbf{Main:} 0.67 ; \textbf{Bias (w/o):} 0.0 (0.77)\\
		& \textbf{H} well his knees were bothering him yeah \\
		& \textbf{P} He was in tip-top condition. \\
		\\
		d) & \textbf{GT:} \textit{C}; \textbf{Main:} 0.39 ; \textbf{Bias (w/o):} 0.1 (0.89)\\
		& \textbf{H} Even us if you needed," said John. \\
		& \textbf{P} He told them not to ask him to lift a finger. \\
		\\
		e) & \textbf{GT:} \textit{N}; \textbf{Main:} 0.72 ; \textbf{Bias (w/o):} 0.1 (0.92)\\
		& \textbf{H} What changed? \\
		& \textbf{P} What was unique? \\

	\end{tabular}
	\caption{Ground Truth (GT) class probabilities of the main and bias models on examples from the MNLI dev set. Examples consist of an~\textbf{H}ypothesis and a~\textbf{P}remise, and valid labels are: \textbf{E}ntailment, \textbf{C}ontradiction, and \textbf{N}eutral. Probabilities of the bias model without reweighting its examples are placed within parenthesis.}
	\label{fig:example}
			
\end{figure} 

We inspected the confident scores of the main and bias (with and without example reweighting) models on MNLI dev set, an excerpt  of which is reported in Figure~\ref{fig:example}. On one hand, we observe that bias models successfully assign high probabilities to samples that can be easily classified via keywords (e.g. "not" in example a) and those with high lexical overlap (example b). On the other hand, we noticed that the bias model is performs undesirably well on some challenging examples like (c) and (d) of Figure~\ref{fig:example}. However, the bias model probability always decreases when we re-weight its examples, which eventually leads to correct prediction of the main model as in example (c) and (e). Interestingly, this observation suggests that training a pure bias model is as important (and challenging) as training the robust one.

\section{Conclusion}

Our key contribution is a framework that jointly identifies biased examples and features in an end-to-end manner. Our approach is geared towards addressing lexico-syntactic bias features for Transformer-based NLU models. Future work involves testing our approach on other tasks such as Question Answering~\cite{rajpurkar2016squad,agrawal2018don}, exploring methods to obtain proxy OOD data for hyper-parameters selection, and making our method hyper-parameter free.

\section*{Acknowledgments}
We thank Mindspore\footnote{\url{https://www.mindspore.cn/}} for the partial support of this work, which is a new deep learning computing framework.
  
\bibliographystyle{acl_natbib}
\bibliography{acl2021}


\end{document}